\documentclass[10pt,twocolumn,letterpaper]{article}

\usepackage{cvpr}
\usepackage{times}
\usepackage{epsfig}
\usepackage{graphicx}
\usepackage{amsmath}
\usepackage{amssymb}

\usepackage{microtype}
\usepackage{graphicx}
\usepackage{booktabs} 
\usepackage{times}
\usepackage{epsfig}
\usepackage{graphicx}
\usepackage{amsmath}
\usepackage{amssymb}
\usepackage{amsthm}
\usepackage{subcaption}
\usepackage{float}
\usepackage{booktabs}

\newcommand{\citet}{\cite}
\newcommand{\citep}{\cite}
\newcommand\supp{\mathop{\rm supp}}

\usepackage[pagebackref=true,breaklinks=true,letterpaper=true,colorlinks,bookmarks=false]{hyperref}

\cvprfinalcopy 

\ifcvprfinal\fi

\begin{document}

%

\title{Towards Robust Image Classification Using Sequential Attention Models}

\author{
Daniel Zoran
\and
Mike Chrzanowski
\and
Po-Sen Huang
\and
Sven Gowal
\and
Alex Mott\and
Pushmeet Kohli\\
\\DeepMind\\
London, UK\\
{\tt\small danielzoran@google.com}
}

\maketitle

\begin{abstract}
In this paper we propose to augment a modern neural-network architecture with an attention model inspired by human perception. Specifically, we adversarially train and analyze a neural model incorporating a human inspired, visual attention component that is guided by a recurrent top-down sequential process. Our experimental evaluation uncovers several notable findings about the robustness and behavior of this new model. First, introducing attention to the model significantly improves adversarial robustness resulting in state-of-the-art ImageNet accuracies under a wide range of random targeted attack strengths. Second, we show that by varying the number of attention steps (glances/fixations) for which the model is unrolled, we are able to make its defense capabilities stronger, even in light of stronger attacks --- resulting in a ``computational race'' between the attacker and the defender. Finally, we show that some of the adversarial examples generated by attacking our model are quite different from conventional adversarial examples --- they contain global, salient and \emph{spatially coherent} structures coming from the target class that would be recognizable even to a human, and work by distracting the attention of the model away from the main object in the original image.

\end{abstract}

\vspace{-0.5em}
\section{Introduction}

\begin{figure}[h!t]
    \centering
    \includegraphics[width=0.99\columnwidth]{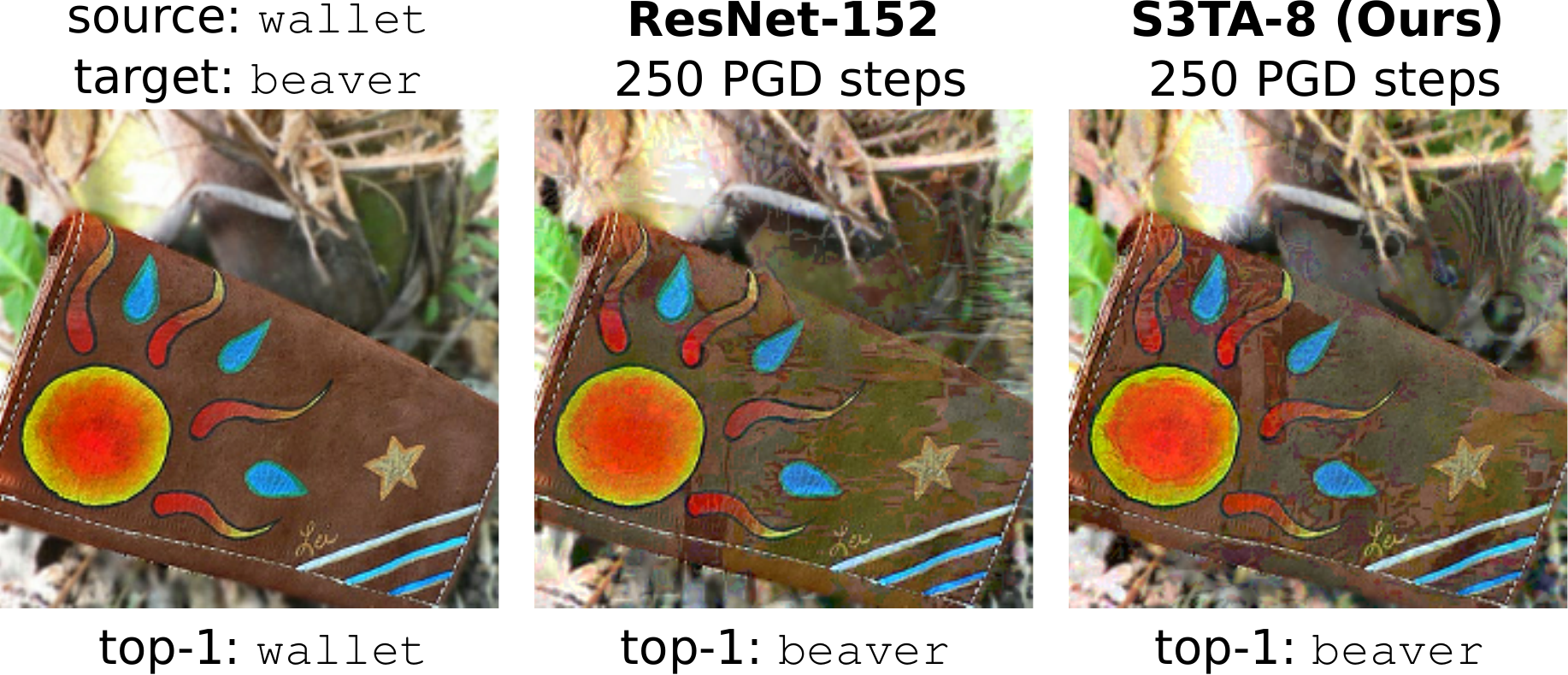}
    \caption{We augment a modern neural network with a sequential top-down attention model for image classification. The model achieves state-of-the-art adversarial robustness against PGD attacks and the resulting adversarial images are often human interpretable. On the left is a source image (label:\texttt{wallet}) --- both an adversarially trained ResNet-152 and our model classify it correctly. In the middle and on the right are adversarial examples produced by a 250 step PGD attack against each model (target class is \texttt{beaver}). Both models fail to defend against the attack and predict the target class as their top-1 output. However, while the attack image for the ResNet contains no visible interpretable structure, the attack image for our model contains a salient and coherent image of a beaver's head (best viewed zoomed in on screen).
    }\vspace{-1em}
    \label{fig:teaser}
\end{figure}

Recent years have seen great advances in the use and application of deep neural network models. From large scale image classification \cite{he2016deep} to speech recognition \cite{hinton2012deep}, the performance of these models has steadily improved, making use of new hardware advances, more memory and better optimization strategies. The leading model paradigm for such tasks, however, has not changed significantly since the original AlexNet paper \cite{alexnet}. Models are still, predominately, built in a purely feed-forward manner, interleaving convolutional layers (often with small kernels with limited support) and simple non-linearities \cite{nair2010rectified}. Recently introduced ResNets \cite{he2016resnet}, which are some of the most powerful models we use currently, have not changed this scenery significantly.

While there is no doubt that these models are very successful in solving some tasks, concerns have been raised about their robustness and reliability \cite{madry2017towards,szegedy2013intriguing}. Small, carefully chosen perturbations to the input, often imperceptible to a human observer, may cause these models to output incorrect predictions with high confidence \citet{szegedy2013intriguing}. These kind of perturbations are called adversarial examples \citep{goodfellow2014explaining,szegedy2013intriguing} and are a subject of ongoing research \cite{athalye2017synthesizing, carlini2017towards, fairdenoising2018}. 

The current paradigm of neural network models has certainly been inspired by the human and primate visual system \citet{Riesenhuber_1999}. Early predecessor models have directly made this connection, and there is a line of work which connects between the activations of such neural network models and the neural activity in brains \citet{Cadieu_2014}. These parallels between models and biological vision systems apply mostly to early vision processing \citet{elsayed2018adversarial} -- and specifically, the feed-forward processing which happens in time-limited scenarios \citet{elsayed2018adversarial}. This has been discussed in several intriguing works, including in the adversarial examples context.

There are, however, some major differences between feed-forward neural network and the primate visual system. The eye in primates has a fovea which samples different regions of the visual input field at different spatial resolutions \citet{Essen1995InformationPS}. Furthermore (and possibly tightly connected to the fovea) the system has a strong attentional bottleneck which has been researched in many different works \citet{roelfsema1998object, baldauf2014neural}. The visual cortex has many feedback and top-down \emph{recurrent} connections \citet{olshausen201320} and it is not purely feed-forward. Additionally, humans don't view images as a static scene, but explore the images in a series of saccades/fixations, collecting and integrating information in the process \cite{liversedge2000saccadic}. This has been postulated to cause humans to report different classification mistakes which are qualitatively different than those of deep neural networks \citet{eckstein2017humans}.

In this work we propose to use a \textbf{s}oft, \textbf{s}equential, \textbf{s}patial, \textbf{t}op-down \textbf{a}ttention mechanism (which we abbreviate as S3TA) \cite{mott2019towards}, drawing inspiration from the primate visual system. While we do not presume this to be a \emph{biologically-plausible} model in any way, we do propose that this model captures some of the \emph{functionality} of the visual cortex, namely the attentional bottleneck and sequential, top-down control. We adversarially train the model on ImageNet images, showing that it has state-of-the-art robustness against adversarial attacks (focusing on Projected Gradient Descent or PGD \cite{kurakin2016adversarial, madry2017towards} attacks). We show that by increasing the number of steps we unroll the model, we are able to better defend against stronger attacks -- resulting in a ``computational race'' between the attacker and the defender. Finally, but importantly, we show that the resulting adversarial examples often (though not always) include global, salient structures which would are perceptible and interpretable by humans (Figure \ref{fig:teaser}). Furthermore, we show that the attack often tries to attract the attention of the model to different parts of the image instead of perturbing the main object in the source image directly.
\section{Related Work}

\textbf{Adversarial training}:
Adversarial training aims to create models that are robust to adversarial attacks.
At their core, techniques such as the ones of \citet{goodfellow2014explaining} and \citet{madry2017towards} find the worst case adversarial examples at each training step (using Fast Gradient Sign Method or PGD attacks) and add them to the training data.
Models created by \citet{madry2017towards} have been shown to be empirically robust on MNIST and CIFAR-10.
\citet{kannan2018adversarial} proposed using Adversarial Logit Pairing (ALP) to encourage the logit predictions of a network for a clean image and its adversarial counterpart to be similar.
However, ALP performs poorly under stronger attacks \cite{engstrom2018evaluating}.
\citet{fairdenoising2018} proposed feature denoising networks together with adversarial training to achieve strong performance on ImageNet \cite{deng2009imagenet}.
Other methods like \cite{guo2017countering} achieve gradient obfuscation more explicitly by adding non-differentiable preprocessing steps.
Although these gradient masking techniques make gradient-based attacks fail, more sophisticated adversaries, such as gradient-free methods \cite{uesato2018adversarial,athalye2018obfuscated}, can circumvent these defenses.

\textbf{Recurrent attention models}:
Attention mechanisms have been widely used in many sequence modeling problems such as question-answering \cite{hermann2015teaching}, machine translation \cite{bahdanau2014neural, vaswani2017attention}, video classification and captioning \cite{shan2017spatiotemporal,li2017mam}, image classification and captioning \cite{mnih2014recurrent, chung2018cram, fu2017look,ablavatski2017enriched,xiao2015application,zheng2017learning,wang2017residual,ba2014multiple,Xu2015ShowAA},
text classification \cite{yang2016hierarchical, shen2016neural}, generative models \cite{parmar2018image, Zhang2018SelfAttentionGA,kosiorek2018sequential}, object tracking \cite{kosiorek2017hierarchical}, and reinforcement learning \cite{choi2017multi}. 
We build out model based on the one introduced in \cite{mott2019towards} and adapt and modify it for ImageNet scale image classification. The model uses a soft key, query, and value type of attention similar to \citet{vaswani2017attention,parmar2018image}. However, instead of using \emph{self}-attention, where the queries come from the input directly, this model uses a top-down source for them generated by a LSTM (see Section \ref{sec:model} for details). Furthermore, the output of the attention model is highly compressed and has no spatial structure other than the one preserved using a spatial basis. This is unlike self-attention where each pixel attends to every other pixel and hence the spatial structure is preserved. Finally, it applies attention sequentially in time similar to \cite{Xu2015ShowAA}, but with a largely different attention mechanism. See Section \ref{sec:model} for full model details.

\textbf{Adversarial robustness with attention}:
There has been some work studying the use of attention to strengthen classifiers against adversarial attacks. \citet{LuoBRPZ15} uses foveation-inspired, manual image cropping to reduce the impact of an adversarial perturbation on ImageNet classification accuracy. \citet{wu2018attention} tries to regularize the activations of a classifier by applying a hard mask generated from an adversarial-perturbed form of the image. The hope is that the use of a mask to occlude important parts of the image builds robustness to perturbations. Recently it has also been shown that ``squeeze and excite'' \citet{hu2018squeeze} self attention can help in classifying ``Natural'' adversarial examples \citet{hendrycks2019nae}. We evaluate our model on this dataset in Section \ref{sec:natural}.
\section{Model}
\label{sec:model}
We base our model on the one proposed by \cite{mott2019towards} for reinforcement learning and we adapt it to ImageNet scale image classification. The model sequentially queries the input, actively attending relevant pieces of spatial information at each time step to refine its estimate of the correct label. The two key components are the \emph{sequential} nature of the model and the \emph{top-down, attentional} bottleneck, both of which we empirically show contribute to its resilience to adversarial attacks.

\begin{figure}
    \centering
    \includegraphics[width=0.95\columnwidth]{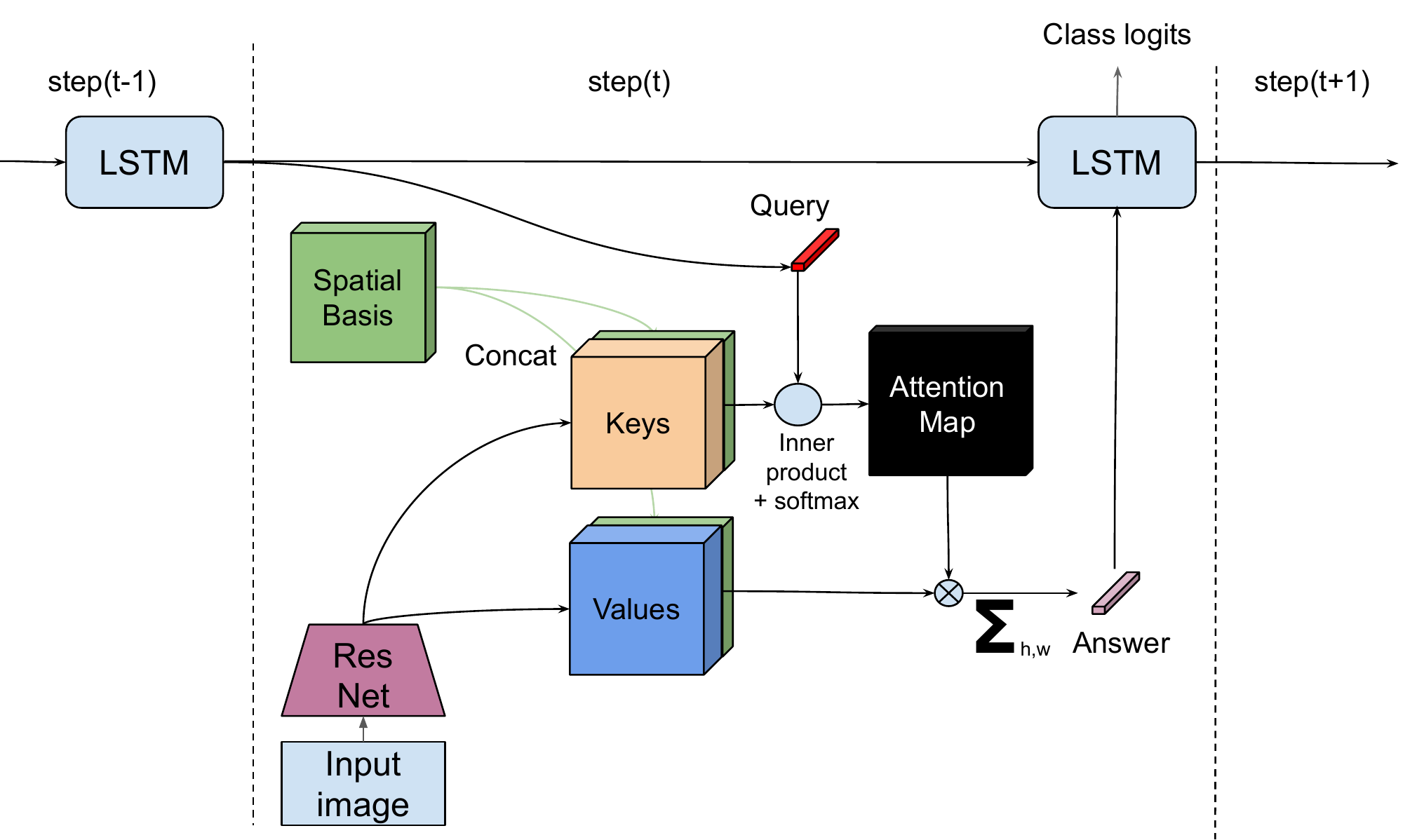}
    \caption{A general view of the sequential top-down attention model. The input image is passed through a ResNet to produce a keys and a values tensor. We concatenate a fixed, predefined spatial basis to both. A query vector is decoded from an LSTM state and an inner product between it and the keys vector is calculated at each spatial location. A softmax is then applied to produce an attention map. This attention map is point-wise multiplied with the values tensor, and the result is summed spatially to produce an answer vector. This answer vector is the input to the LSTM at this time step. The LSTM output is then decoded into class logits to produce the output of the classifier. More than one query vector can be produced at each time step (resulting in a matching number of answer vectors).}
    \label{fig:model}
\end{figure}

We briefly highlight the important components of the model, illustrated in Figure \ref{fig:model}. For full details, we refer the reader to \cite{mott2019towards} and the supplementary material.
The model starts by passing the input image through a ``vision'' net --- a convolutional neural net (here we use a modified ResNet-152, see below). We use the same input image for all time steps, so the output of the ResNet needs to be calculated only once. The resulting output tensor is then split along the channel dimension to produce a \emph{keys} tensor and a \emph{values} tensor. To both of these tensors, we concatenate a fixed \emph{spatial basis} tensor which encodes spatial locations using a Fourier representation. This spatial basis is important because our attentional bottleneck sums over space causing the spatial structure of these tensors to disappear and this basis allows passing on spatial location information.

We unroll the top-down controller for several compute steps, attending the input at each step and processing the answer through the controller to produce the output (or next state). The Top-Down controller is an LSTM core \cite{Hochreiter:1997:LSM:1246443.1246450} whose previous state is decoded through an a ``query network'', an MLP, into one or more \emph{query} vectors. Each query vector has the same number of channels as the keys tensor plus the number of channels in the spatial basis. We take the inner product between the query vector and key and spatial basis tensor at each spatial location, resulting in a single channel map of attention logits. We pass these through a spatial softmax to produce the \emph{attention map} for this query. The resulting attention map is then point-wise multiplied with the values tensor (and the spatial basis). Note that a single map is used for \emph{all} channels, we note the importance of this below. The multiplied value tensor is summed across the \emph{spatial} dimensions to produce an \emph{answer} vector, one for each query. These answers are fed into the LSTM as the input for the current time step (concatenating them if more than one is used).
Finally, the output from the \emph{last} LSTM output is decoded into class logits through an MLP. The cross-entropy loss w.r.t to the ground truth class is calculated with this output. The initial state of the LSTM is also learned. Since the model is fully differentiable, we train it end-to-end, including the ResNet, with adversarial training (as described in Section \ref{adversarial_training}) and \ref{sec:experiments}.

Several important points regarding our version of the model in this context:
\begin{itemize}
    \item The attention bottleneck makes the decision of the model depend on potentially large extents of the image. This can be due to the shape of attention map at every time step, as well as the fact that these maps can change considerably between time steps. This should cause a \emph{local} adversarial perturbation \cite{universal_adv_perturbations} to be less effective. We discuss this in Section \ref{sec:analysis} and show that indeed, we often observe that global perturbations are required by the attacker for the attack to succeed.
    \item Following the last point, the fact that the attention map has a single channel which modulates all value channels \emph{together} constraints the content of these channels to be spatially coherent. In a regular ResNet architecture the last block output is read with a average pooling done independently on each channel - this allows the network to lose spatial structure by the time information reaches this last layer.
    \item In order to make the spatial element, and hence the effect of the attention bottleneck more pronounced we modify the ResNet architecture to make the final output have larger spatial dimensions. This is done by changing the strides to 1 in all but the second residual block. For ImageNet input ($224\times224$ pixels) the resulting map is $28\times28$ pixels large (as opposed to $7\times7$ in a regular ResNet).
    \item  The top-down nature of the attention mechanism is such that the queries come from the state of the LSTM and not from the input. Hence, the model can actively select relevant information depending on its internal state, rather than just the input. This allows the model to take its own uncertainty, for example, into account, when querying the image and producing the output.
    \item The sequential nature of the model allows for increasing computational capacity without changing the number of parameters. We demonstrate that this helps with robustness in Section \ref{sec:experiments}.
\end{itemize}
\section{Adversarial Risk}

We define adversarial risk in the context of supervised learning formally in this section.
Given a model $m_{\theta}$ with parameters $\theta$, we want to minimize the loss $\ell$ on inputs $x$ and labels $y$ sampled from the data distribution $D$. 
Formally, the objective is to minimize the \emph{expected} risk:
$
    \mathop{\mathbb{E}}_{(x,y)\sim D} \ell(m_{\theta}(x), y).
$ 
Empirically, we optimize the empirical risk on a finite training set and estimate the expected risk over a held-out test set using the average loss.

As pointed out in \cite{uesato2018adversarial}, models with low expected risk may still perform poorly on any data points. In situations where a single catastrophic failure is not allowable, the empirical risk estimate may be problematic. 
Hence, we also need to consider the \emph{worst-case} risk for the desired robust models:
$\sup_{(x, y) \in \supp{D}} \ell(m_{\theta}(x), y)$,
where $\supp{D}$ denotes the support of $D$.
In practice, computing the supremum over the input space is intractable as the search space is exponentially large in the dimension of $x$.
We can instead use the local \textit{adversarial risk}, as a proxy for the worst-case risk:
\begin{equation}
\label{eq:adv_risk}
    \mathop{\mathbb{E}}_{(x,y)\sim D } \left[ \sup_{x' \in N_\epsilon(x)} \ell(m_{\theta}(x'), y) \right],
\end{equation}
where the neighborhood $N_\epsilon(x)$ denotes a set of points in $\supp(D)$ within a fixed distance $\epsilon > 0$ of $x$, measured by a given metric.
The adversarial risk enables us to approximate the worst-case risk in a tractable way. For example, we can use off-the-shelf optimization algorithms (such as PGD \cite{kurakin2016adversarial, madry2017towards}) to find the supremum over the neighborhood $N_\epsilon(x)$.

In this paper, we consider the specification that the image predictions should remain the same within an $\ell_\infty$-ball of an image $x$, where an allowable maximum perturbation is $\epsilon=16/255$,  relative to the pixel intensity scaled between 0 and 1.   

Specifically, we focus on the ImageNet dataset \cite{deng2009imagenet} and we primarily consider the targeted PGD attack as the threat model,
where the targeted class is selected uniformly at random, following \cite{athalye2018obfuscated, kannan2018adversarial, fairdenoising2018},
given that the untargeted attacks can result in less meaningful comparisons (e.g., misclassification of very similar dog breeds) on ImageNet.

\subsection{Adversarial Training}
\label{adversarial_training}

To train models that are robust to adversarial attacks, we follow the adversarial training approach by \cite{madry2017towards} and more recently \cite{fairdenoising2018}.

Following the adversarial risk in Eq. \eqref{eq:adv_risk}, we want to minimize the following saddle point problem:
\begin{equation}
\label{eq:minmax}
	\min_\theta \rho(\theta),~\text{where}\quad \rho(\theta) =
    \mathop{\mathbb{E}}_{(x,y)\sim D }\left[\max_{x' \in N_\epsilon(x)}
    \ell(m_{\theta}(x'), y)\right] \; .
\end{equation}
where the inner maximization problem is to find an adversarial perturbation of $x$ that can maximize the loss; the outer minimization problem aims to update model parameters such that the adversarial risk $\rho(\theta)$ is minimized.

In our experiments, we approximate the solution to the inner maximization problem with PGD. Specifically, we perform PGD on the cross entropy loss described using iterative signed gradients as in \citet{kurakin2016adversarial, fairdenoising2018}.
During training, we use targeted PGD attacks, where the targeted class is selected at random uniformly, following \cite{kannan2018adversarial, fairdenoising2018}.

 \subsection{Adversarial Evaluation}
In this paper, we use the PGD attack to evaluate the model, which is regarded as a strong attack\footnote{Note that PGD is not necessarily the most suited attack for sequential models, but for lack of better alternative we use it with large number of steps.} in the community and several published papers use this as their benchmark. 

In cases where we cannot take analytic gradients, or where they are not useful, we can approximate the gradients using gradient-free optimization. 
The use of gradient-free methods lets us verify whether robustness stems from gradient obfuscation by the model architecture.
In this work, we use the SPSA algorithm \cite{spall1992multivariate}, which is well-suited for high-dimensional optimization problems, even in the case of noisy objectives. 
We use the SPSA formulation in \citet{uesato2018adversarial} to generate adversarial attacks. 
In the SPSA algorithm, it first samples a batch of $n$ samples from a Rademacher distribution (i.e., Bernoulli $\pm 1$), namely, $v_1, \ldots, v_n \in \{1, -1\}^D$.~
Then, the SPSA algorithm approximates the gradient with finite difference estimates in random directions. Specifically, for the $i$-th sample, the estimated gradient $g_i$ is calculated as follows:
\[
g_i = \frac{f(x_t+\delta v_i) - f(x_t - \delta v_i)}{2 \delta v_i}
\]
where $\delta$ is the perturbation size, $x_t$ is the perturbed image at the $t$-th iteration, and $f$ is the model to be evaluated. 
Finally, SPSA aggregates the estimated gradient and performs projected gradient descent on the input $x_t$. 
The whole process iterates for a predefined number of iterations.
\section{Experiments}
\label{sec:experiments}

In this section, we empirically study the robustness of S3TA on the ImageNet dataset \cite{deng2009imagenet}. For convenience, we denote S3TA-$k$ to be a S3TA model that is unrolled for $k$ time steps and evaluate S3TA-2, S3TA-4, S3TA-8, and S3TA-16.

We follow the training procedure used in \citet{fairdenoising2018}, including learning rate schedule, label smoothing, attack type during training and evaluation procedure. We find that training with a somewhat lower learning rate (0.2 initial learning rate) and a smaller batch size (1024) is more stable for our model. Training S3TA-16 is more challenging than the other models due to the length of the unroll. In order to train it we start by reading off the output from the 4th step for the first 35 epochs, 8th step for the next 35 epochs and 16th step for the rest of training. All models are trained for 120 epochs. We train the model on 128  Google Cloud TPU v3 cores. Training takes between 42 and 70 hours, depending on the number of unroll steps. We use a ResNet-152 as the vision-net of the model (see Section \ref{sec:model}) setting all strides to 1 other than the second residual block. This results in larger spatial support for the ResNet output ($28\times28$ pixels) The recurrent core is an LSTM with 1024 hidden units, the query network and the output MLP are both with a single hidden layer of 1024 units. All the activation units used are ReLUs. The attention model uses 4 attention query heads in all experiments here.

\subsection{Random targeted attacks}
\begin{figure}[t]
    \centering
    \includegraphics[width=0.4\textwidth]{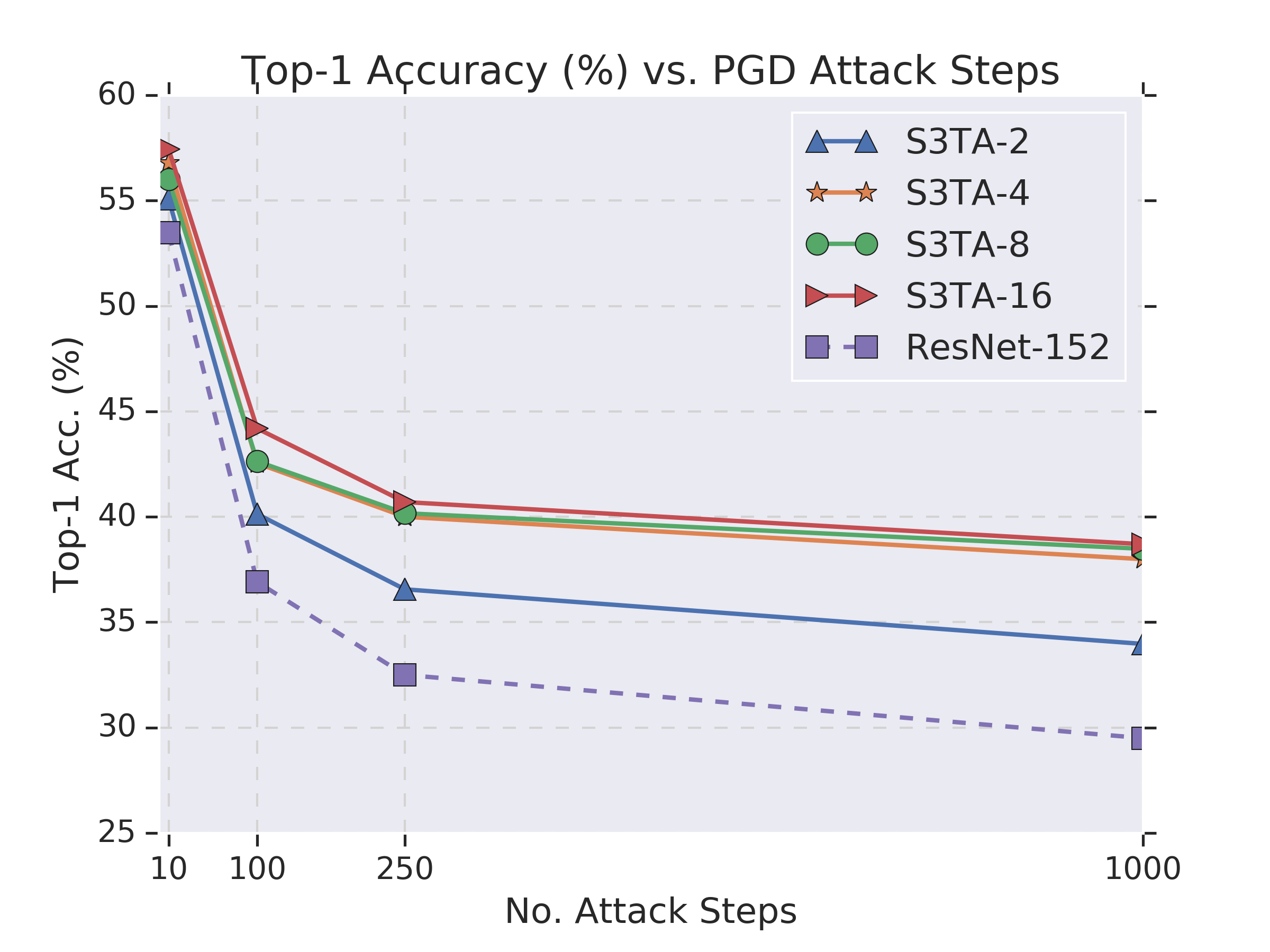}\vspace{-1em}
    \caption{S3TA-2, 4, 8 and 16 vs. ResNet-152 top 1 accuracy on the ImageNet test set.. All models were adversarially trained with 10 PGD steps. Note how the introduction of the attention model significantly improve performance even with 2 attention steps, and that adding more steps (S3TA-16) improves performance further: a S3TA-16 model is more robust at a 1000 attack steps than a ResNet-152 model at a 100 attack steps.}
    \label{fig:top1_10steps}
\end{figure}

The first set of models was adversarially trained with 10 PGD steps. These are generally weaker models than models trained with 30 PGD steps (see below) but they take less time and resources to train. Figure \ref{fig:top1_10steps} shows the top-1 accuracy of these models for the ImageNet test dataset under a wide range of random targeted PGD attack strengths compared to a ResNet-152 baseline (also trained with 10 PGD steps during adversarial training).
With only 2 steps of attention the weakest model here, S3TA-2, only has a chance to send two queries, one before it even sees the image, and one after processing the answers from the first step. This puts emphasis on the \emph{attention bottleneck} itself rather than the sequential nature of the model. As can be seen, the bottleneck itself already allows to model to improve significantly upon the ResNet-152 baseline.

\begin{figure}[t]
    \centering
    \includegraphics[width=0.4\textwidth]{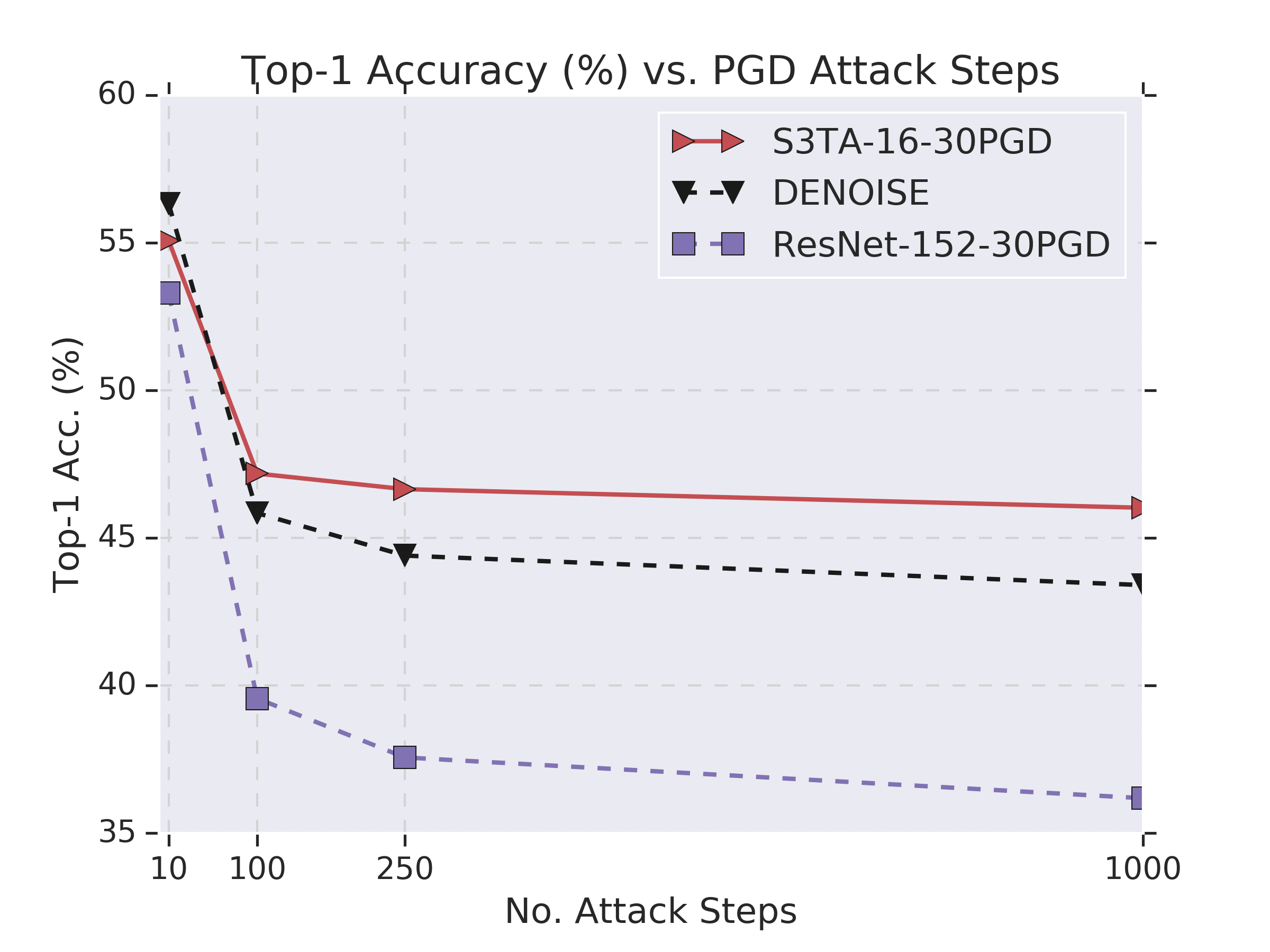}\vspace{-1em}
    \caption{A S3TA-16 model adversarially trained with 30 PGD steps vs. ResNet-152 (30 steps) and DENOISE \cite{fairdenoising2018} top 1 accuracy on the ImageNet test set. DENOISE is the current state-of-the-art on ImageNet and as can be seen S3TA-16 performs significantly better than both models, setting a new state-of-the-art.}
    \label{fig:top1_30steps}
\end{figure}

By increasing the number of attention steps we can improve adversarial accuracy even further: unrolling for 16 steps (S3TA-16) significantly improves robustness - a S3TA-16 model is more robust against a 1000 PGD attack steps than a ResNet-152 model is against a 100 attack steps. In fact, a S3TA-16 model trained with 10 PGD steps during adversarial training is more robust than a ResNet-152 \emph{trained with 30 PGD steps} (see Figure \ref{fig:top1_30steps}). This shows that there is a kind of ``computational race'' here between the strength of the attack and the number of compute steps we allow the model to have. More computation steps for the model mean better defense against stronger attacks. Going beyond a 1000 attack steps does not change the picture as most models saturate close to their 1000 step performance. Full results, including attack success rates, and nominal accuracies, can be found in Table \ref{tab:full_results} and the supplementary material.

We now turn to compare models adversarially trained with 30 PGD steps. These models are much stronger generally and achieve good robustness results across a wide range of attack strengths, but require a great deal of resources and time to train. Figure \ref{fig:top1_30steps} shows the top-1 accuracy of a S3TA-16-30 model (``-30'' denotes 30 PGD steps during training) vs. a ResNet-152 model and DENOISE \cite{fairdenoising2018}, the latter being the current state-of-the-art in adversarial robustness. As can be seen, S3TA-16 comfortably outperform both models, setting a new state-of-the-art for random targeted attacks. 

\begin{figure}[t]
    \centering
    \includegraphics[width=0.4\textwidth]{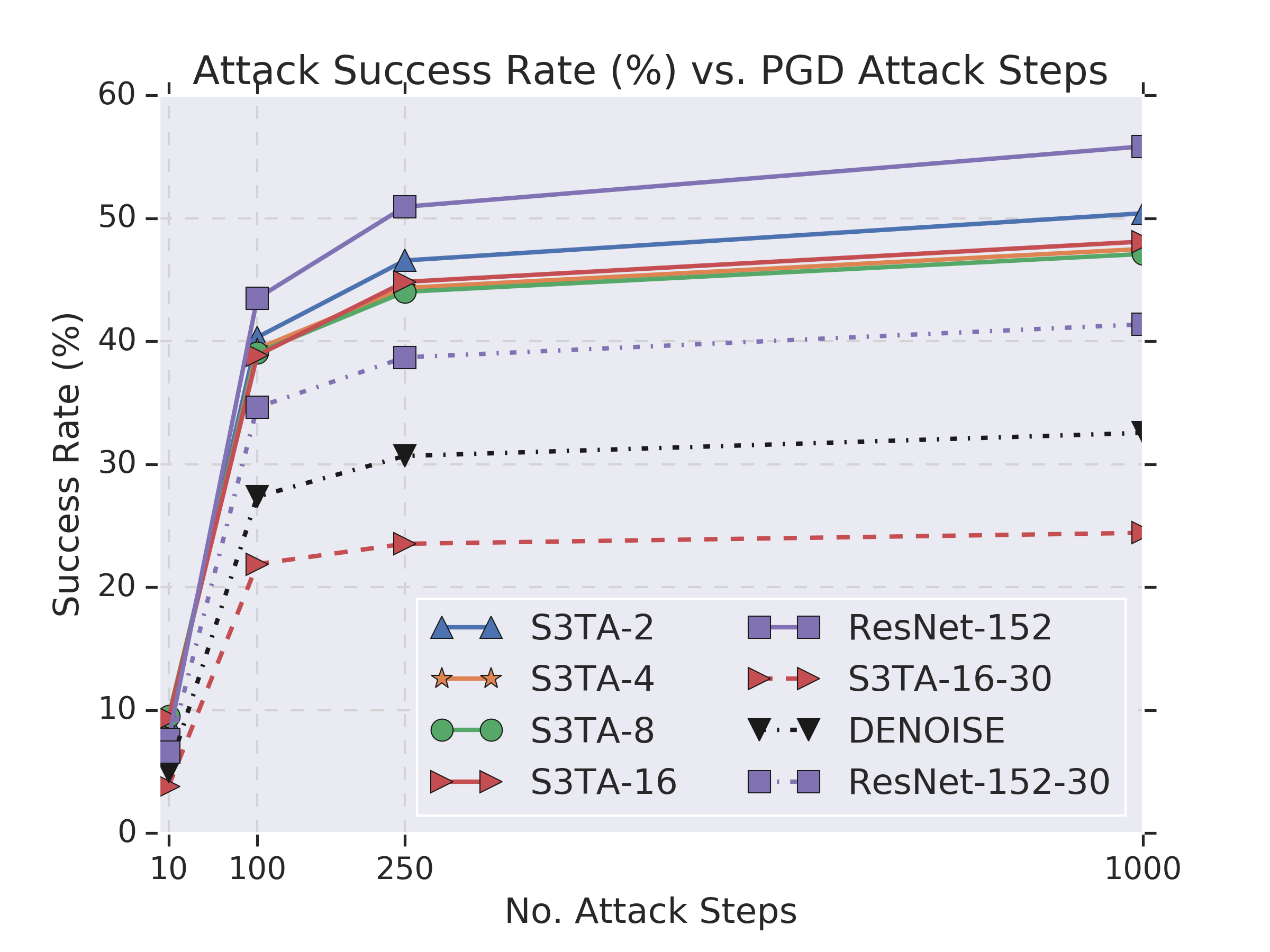}\vspace{-1em}
    \caption{Attack success rates for all models presented (lower is better). The main effects observed for top 1 accuracy hold here: more attention steps lower the attack success rate and more PGD steps during training help reduce it even further. S3TA-16-30 clearly has the lowest attack success rates, about 25\% lower than DENOISE while nominal accuracy is similar (see \ref{tab:full_results}) }
    \label{fig:success_rates}
\end{figure}

Figure \ref{fig:success_rates} shows the attack success rates for all models discussed so far. When evaluating defense strategies, measuring attack success rates makes sense when nominal accuracies of models are high and comparable. For all models presented here this is true (see Table \ref{tab:full_results}). Note that the results hold for this measure as well - more attention steps help reduce attack success rates, and more PGD steps during training help. The success rate of attacks against S3TA-16-30 is about 25\% lower than that of DENOISE (lower is better).

\begin{table*}[]
\centering
\begin{tabular}{@{}ll|l|l|l|l|l|l|l|l|@{}}
\cmidrule(l){3-10}
\multicolumn{1}{c}{\textbf{}} & \multicolumn{1}{c|}{\textbf{}} & \multicolumn{2}{c|}{10 steps} & \multicolumn{2}{c|}{100 steps} & \multicolumn{2}{c|}{250 steps} & \multicolumn{2}{c|}{1000 steps} \\ \cmidrule(l){3-10} 
\multicolumn{1}{l|}{\textbf{Model}} & \textbf{\begin{tabular}[c]{@{}l@{}}Nominal\\ Accuracy\end{tabular}} & \multicolumn{1}{c|}{\begin{tabular}[c]{@{}c@{}}Top-1\end{tabular}} & \multicolumn{1}{c|}{\begin{tabular}[c]{@{}c@{}}Attack\\ Success\end{tabular}} & \multicolumn{1}{c|}{\begin{tabular}[c]{@{}c@{}}Top-1\end{tabular}} & \multicolumn{1}{c|}{\begin{tabular}[c]{@{}c@{}}Attack\\ Success\end{tabular}} & \multicolumn{1}{c|}{\begin{tabular}[c]{@{}c@{}}Top-1\end{tabular}} & \multicolumn{1}{c|}{\begin{tabular}[c]{@{}c@{}}Attack\\ Success\end{tabular}} & \multicolumn{1}{c|}{\begin{tabular}[c]{@{}c@{}}Top-1\end{tabular}} & \multicolumn{1}{c|}{\begin{tabular}[c]{@{}c@{}}Attack\\ Success\end{tabular}} \\ \midrule
\multicolumn{1}{l|}{\textbf{ResNet-152}} & 70.66\% & 53.48\% & 7.63\% & 36.91\% & 43.48\% & 32.50\% & 50.93\% & 29.5\% & 55.84\% \\
\multicolumn{1}{l|}{\textbf{S3TA-2}} & 72.30\% & 55.08\% & 9.06\% & 40.13\% & 40.31\% & 36.56\% & 46.56\% & 33.97\% & 50.40\% \\
\multicolumn{1}{l|}{\textbf{S3TA-4}} & 72.48\% & 56.78\% & 9.10\% & 42.54\% & 39.37\% & 40.00\% & 44.33\% & 37.99\% & 47.50\% \\
\multicolumn{1}{l|}{\textbf{S3TA-8}} & 72.14\% & 56.02\% & 9.50\% & 42.63\% & 39.06\% & 40.17\% & 44.01\% & 38.48\% & 47.09\% \\
\multicolumn{1}{l|}{\textbf{S3TA-16}} & \textbf{72.54\%} & \textbf{57.45\%} & 9.33\% & 44.19\% & 38.83\% & 40.71\% & 44.82\% & 38.70\% & 48.16\% \\ \midrule
\multicolumn{1}{l|}{\textbf{ResNet-152-30}} & 63.62\% & 53.30\% & 6.56\% & 39.56\% & 34.62\% & 37.56\% & 38.68\% & 36.18\% & 41.37\% \\
\multicolumn{1}{l|}{\textbf{DENOISE}} & 66.02\% & 56.33\% & 5.00\% & 45.84\% & 27.36\% & 44.19\% & 30.66\% & 43.39\% & 32.54\% \\
\multicolumn{1}{l|}{\textbf{S3TA-16-30}} & 64.55\% & 55.08\% & \textbf{3.79\%} & \textbf{47.18\%} & \textbf{21.87\%} & \textbf{46.65\%} & \textbf{23.52\%} & \textbf{46.11\%} & \textbf{24.91\%}
\end{tabular}
\caption{Full results for all models on random targeted PGD attacks with the ImageNet test set, with different number of attack steps. Bottom three rows are models trained with 30 PGD steps, the rest were trained with 10~PGD steps.}\vspace{-1em}
\label{tab:full_results}
\end{table*}

\subsection{Untargeted and gradient-free attacks}
Most robustness measures in the literature are for targeted, gradient based attacks. 
However, a model that is only robust against targeted attacks is weaker than one robust against untargeted attacks \cite{engstrom2018evaluating}. 
In Table \ref{tab:untargeted}, we report results for untargeted attacks using 200 PGD steps for S3TA-16-30 vs. ResNet-152, DENOISE and LLR \cite{qin2019adversarial}. 
Our model is very competitive in this setting, both for $\epsilon=4/255$ and $\epsilon=16/255$.

\begin{table}[!ht]
    \centering
    \begin{tabular}{l|c|c|c}
         {\bf Model} & {\bf Top-1 $\epsilon=4/255$} & {\bf Top-1 $\epsilon=16/255$} \\
         \hline
         ResNet-152 & 39.7\% & 6.3\% \\
         DENOISE~\cite{fairdenoising2018} & 38.9\% & 7.5\% \\
         LLR~\cite{qin2019adversarial} &  \bf{47.0\%} & 6.1\% \\
         S3TA-16  & 46.75\% & \bf{9.8\%} \\
    \end{tabular}
    \caption{Top-1 accuracy under untargeted attacks at 200 PGD steps. As can be seen, our model is very competitive with existing methods though not optimized for this particular attack method.
    }\vspace{-1em}
    \label{tab:untargeted}
\end{table}

We also explore gradient-free methods to make sure the model does not obfuscate gradients \cite{uesato2018adversarial, obfuscated-gradients}.
Specifically, we use random targeted SPSA \cite{uesato2018adversarial} with a batch size of 4096 and 100 iterations under $\epsilon=16/255$ for the gradient-free attack. We use iterative signed gradients \citet{kurakin2016adversarial, fairdenoising2018} with gradients estimated by SPSA. 
Results on a subset of 1000 randomly-chosen images can be seen in Table \ref{tab:spsa}. 
We can observe that SPSA does not lower accuracy compared to gradient-based attacks. 
This provides an additional evidence that the model's strong performance is not due to gradient masking. Given SPSA's adversarial accuracy is weaker (that is, all models defend better than with gradient-based methods) the performance difference between the models is not very informative.
\begin{table}[!ht]
    \centering
    \begin{tabular}{l|c|c|}
         {\bf Model} & {\bf Top-1} & {\bf Attack Success}  \\
         \hline
         ResNet-152 & 61.90\% & 2.20\% \\
         DENOISE~\cite{fairdenoising2018} & 63.70\% & 1.90\%  \\
         S3TA-16  & 59.60\% &  1.90\%\\
    \end{tabular}
    \caption{Top-1 accuracy under random targeted SPSA attacks (batch size of 4096 and 100 iterations). SPSA is a gradient free method which provides evidence whether gradients are obfuscated. As can be seen all models perform similarly, considering they all defend better here than the corresponding gradient based attack (making the actual reported number less informative).
    }\vspace{-1em}
    \label{tab:spsa}
\end{table}

\subsection{Loss landscapes}
\begin{figure}[b]
    \centering
    \includegraphics[width=.22\textwidth]{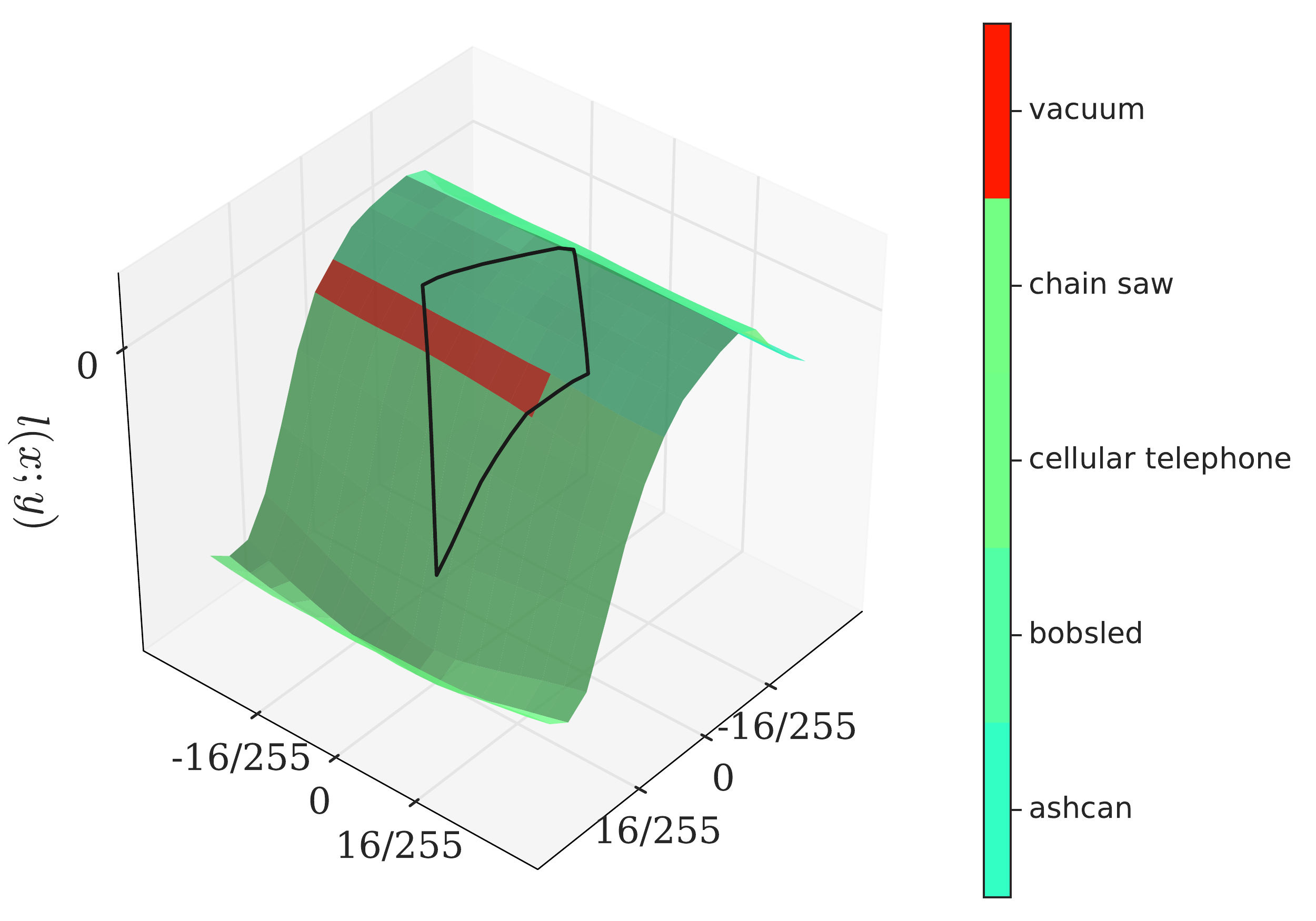}
    \includegraphics[width=.22\textwidth]{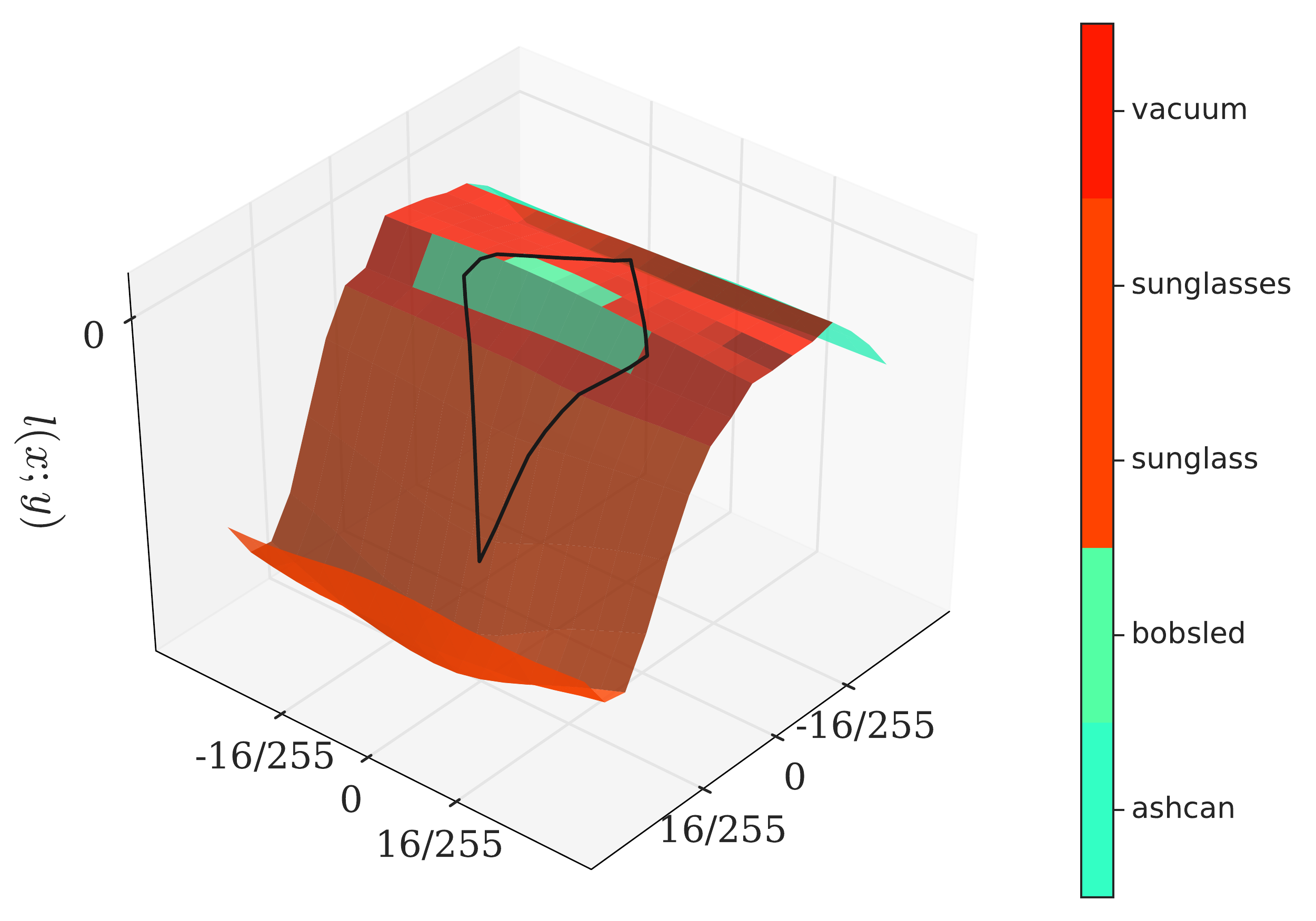}
    \caption{Loss landscapes for S3TA-4 (left) and S3TA-16 (right). The surfaces are approximately linear meaning there is no significant gradient obfuscation.}
    \label{fig:loss}
\end{figure}

Another way of making sure gradients are not obfuscated is by visualizing the loss landscapes \cite{qin2019adversarial, uat2019}.
Figure \ref{fig:loss} shows the top-view of the loss landscape for S3TA-4 and S3TA-16.
To visualize the loss landscapes, we change the input along a linear space defined by the worse perturbations found by PGD and a random direction.
The $u$ and $v$ axes represent the magnitude of the perturbation added in each of these directions respectively and the $z$ axis represents the loss.
For both panels, the diamond-shape stands for the projected $L_\infty$~ball of size $\epsilon = 16/255$ around the nominal image.
We can observe that both loss landscapes are rather smooth, which provides an additional evidence that the strong performance is not because of gradient obfuscation.

\subsection{Natural adversarial examples}
\label{sec:natural}
\begin{figure}[t]
    \centering
    \includegraphics[width=0.4\textwidth]{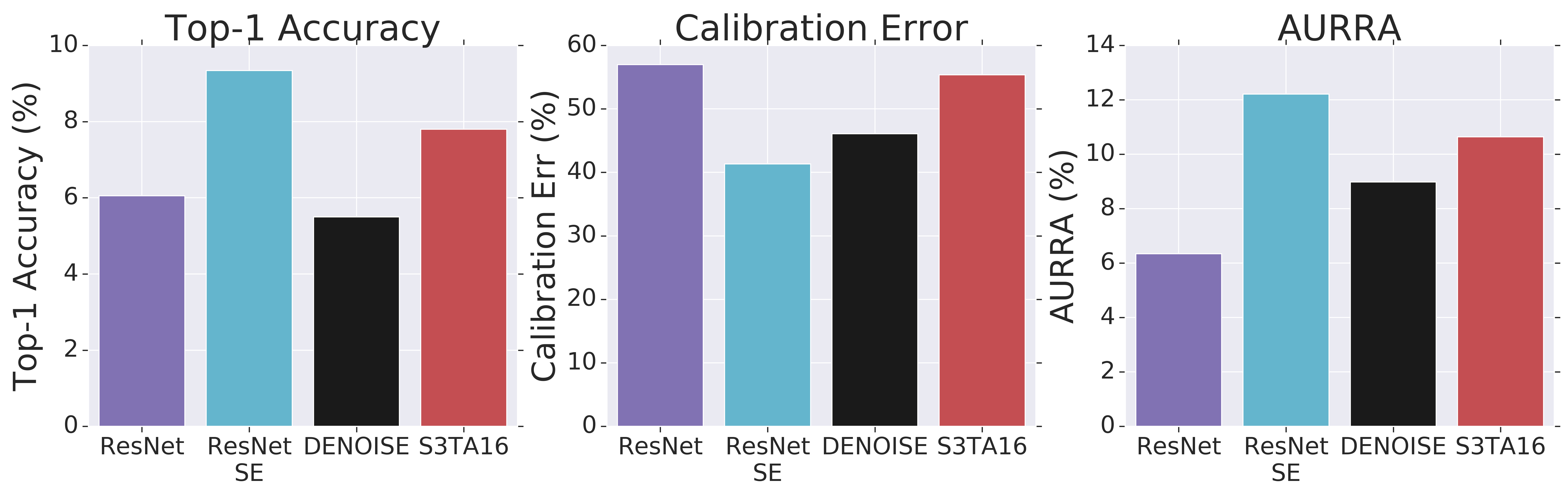}\vspace{-1em}
    \caption{Results of our model on the ``Natural Adversarial Examples'' dataset. Our model achieves better top-1 accuracy, and better AURRA (see text for details) than both a ResNet-152 and DENOISE. Squeeze and excite outperforms our model in all measures.}
    \label{fig:nae}
\end{figure}

A recent interesting dataset is ``Natural adversarial examples'' \citet{hendrycks2019nae}. This curated dataset is composed of \emph{natural} images of a subset of 200 classes from ImageNet. These images are chosen such that they cause modern image classifier to misclassify an image with high confidence, even though no actual modification to the image is done. The images often contain objects in unusual locations, photographed from unusual angles or occluded or corrupted in a variety of ways. We compare a S3TA-16 model to \texttt{DENOISE}, the ResNet baseline and the ``Squeeze and excite''  \citet{hu2018squeeze} (ResNet+SE) variant reported in the original paper. Figure \ref{fig:nae} shows the results, using the measures used in the paper: Top-1 accuracy, Calibration error which measures the difference between the confidence of each model and its actual error rate, and AURRA which allows calculation of accuracy while giving classifiers an opportunity to abstain if they are not confident in their prediction.
\section{Analysis}
\label{sec:analysis}
\begin{figure*}[h]
    \centering
    \includegraphics[width=0.99\textwidth]{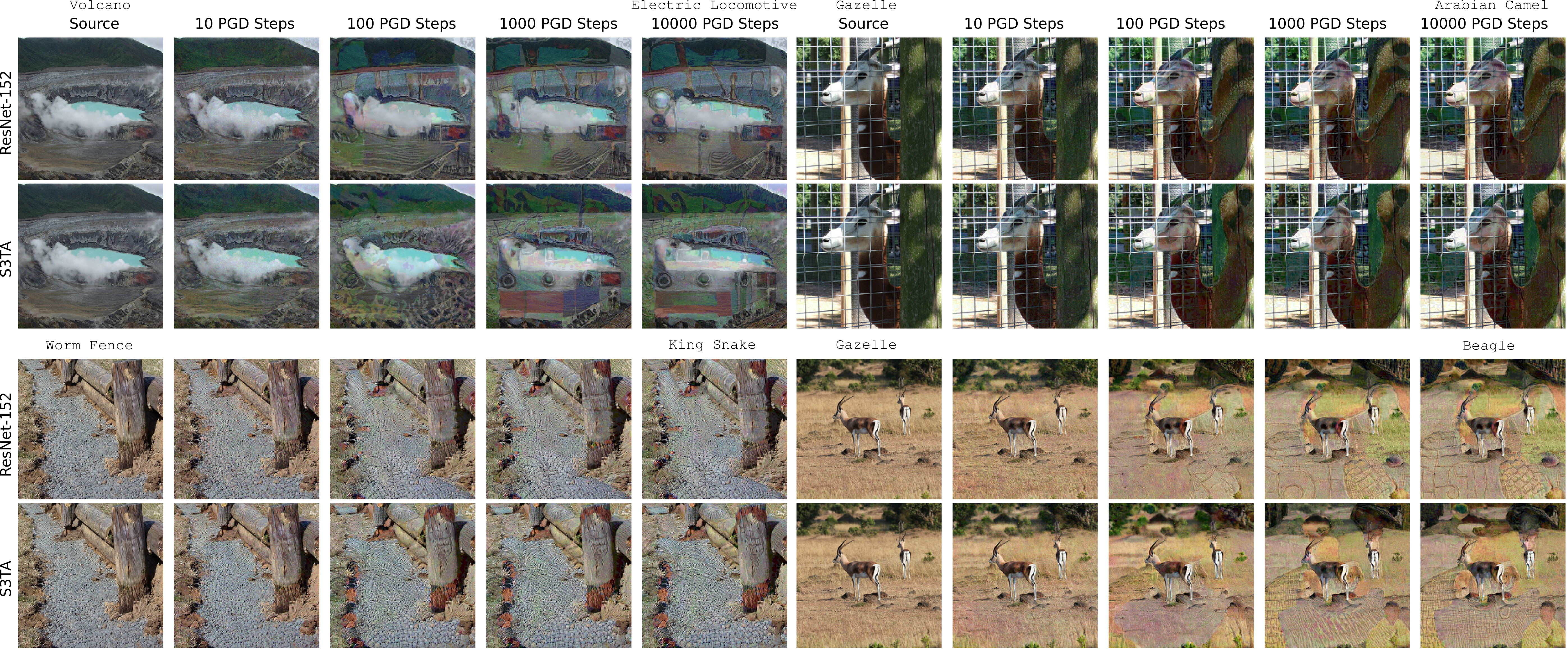}\vspace{-1.0em}
    \caption{Adversarial images generated by PGD to attack the ResNet-152 and S3TA-4 model. The source label is depicted on the left, the target label on the right. Source images are the leftmost column. We generate examples for 10, 100, 1000 and 10,000 PGD steps. Note how for the examples shown here, the perturbations are quite visible for both models (not surprising in light of the strength of the attack). However, the perturbations generated to attack the ResNet model are mostly local, comprising at best of disconnected visible features related to the target class. On the other hand, the examples generated to attack S3TA contain global, coherent and human interpretable structure. Note the 3D structure and spatial extent of the locomotive (top-left), the coherency of the king snake on the ground (bottom-left), the camel head on the bark of the tree (top-right) and the beagle and the man appearing (bottom-right). These structures appear mostly when there are already existing features in the image which can be used to form the target class, and the model uses them in a global, coherent manner. Images are best viewed on screen and zoomed in.}
    \label{fig:examples}
\end{figure*}

We have shown that the sequential attention model improves robustness against a variety of attacks and attack strengths. Furthermore, we have seen that we can increase accuracy and defend better against stronger attacks by unrolling the model for more time steps. We now turn to analyze some of the properties of the resulting attack images and strategies.

Figure \ref{fig:examples} shows several examples of generated adversarial examples for different attack strengths for an adversarially trained S3TA model (with 4 unrolling steps) and an adversarially trained ResNet-152. We observe that often (but certainly not always, see below) the generated image contains salient structures related to the target class. However, while the nature of these perturbations is local at best for the ResNet examples, for S3TA \emph{global, coherent and human intrepretable} structures appear. This sheds some light to the internal reasoning process of our model, hinting that it reasons globally, across space, in a coherent manner. It's important to note that in many cases the adversarial examples do not appear to contain any salient structure (even with many attack steps). They do appear much more often midway through training the model, while the model is already a very good classifier but still not at the peak of its robustness. Towards the end of training it seems that it is harder to generate these, possibly as part of the defense strategy the model learns. See the supplementary material for some examples generated midway through training, as well as more visible and invisible perturbation images. Understanding under what circumstances these examples appear is left for future research.

\subsection{Distracting the Attention}
\begin{figure*}[h]
    \centering
    \includegraphics[width=0.9\textwidth]{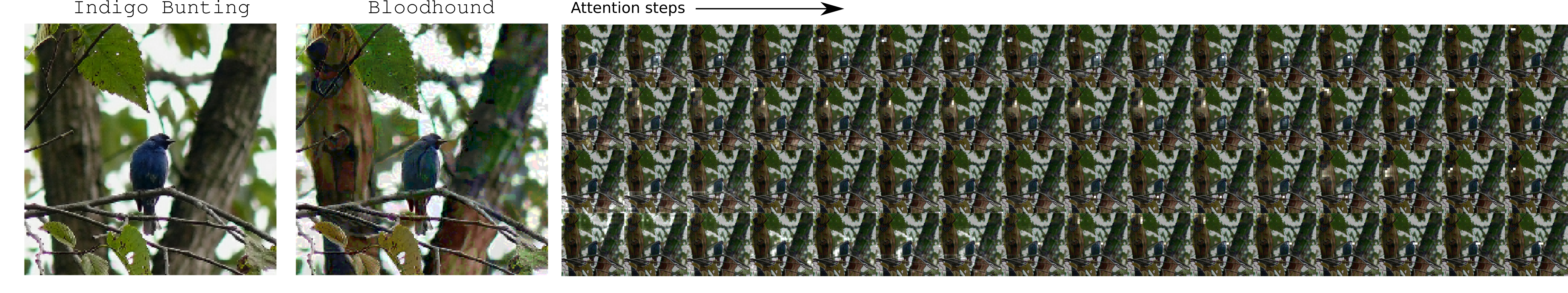}\vspace{-1.0em}
    \caption{Attacks attract attention away from the main object. Here we see the source image (left) attack image (target class \texttt{bloodhound}) and the 4 attention heads of the model as they unroll through 16 steps. Some of the heads still attend the main object from the source image, however some of the heads are attracted away towards a group of branches in the background. Upon close inspection it does seem that these branches resemble a bloodhound dog in the attack image. Though this structure is not very salient to humans, it is enough to attract the attention away and cause the model to mislabel the image. Best viewed on screen and zoomed in.}
    \label{fig:attention}
\end{figure*}
Since attention is an integral part in our model, we can see whether it plays a role when the network is attacked and mislabels an image. We can visualize the attention maps generated at each time step and see how the attention is used under different attack scenarios. Figure \ref{fig:attention} shows such attention maps for an image used to attack a S3TA-16 model. Attention is superimposed over original image - highlighted areas are more attended than dark areas. As can be seen, the attack can create stimuli which attract the some the attention heads away from the main object in the image, in this case towards something that slightly resembles the target class in the background.

\section{Conclusion}
In this paper, we have shown that a recurrent attention model inspired by the primate visual system is able to achieve state-of-the-art robustness against random target adversarial attacks. Allowing for more attention steps improves accuracy under stronger attacks. We show that the resulting adversarial examples often (but not always) contain global structures which are visible and interpretable to a human observer. 

Why is it that global structures arise when attacking a model like this? We postulate that there are two contributing factors. The attention mechanism pools data from large parts of the image, which means that the gradients propagate quickly across the whole of the image, and not just locally. Furthermore, because the the model is unrolled for several steps, more parts of the image may be potentially attended to and thus gradients may propagate there. We see evidence for this in the fact the often the attacker attracts the attention away from the main object in the image, hinting that the attention plays a crucial role in the attack strategy.

There is still a lot of work to be done to achieve adversarial robustness in complex datasets. Even models like the proposed one often fail when the attacker is strong enough, and performance is still quite low compared to nominal accuracies, but at some point we may ask --- if an image has been perturbed enough such that it does not resemble the original image and looks like another image coming from the target class, is it still a valid adversarial perturbation? Models like the one presented here may allow us to reach that frontier in the future.
{\small
\bibliographystyle{ieee_fullname}
\bibliography{main}
}
\newpage
\appendix
\section{Model details}
\begin{figure*}[h!]
    \centering
    \includegraphics[width=0.95\textwidth]{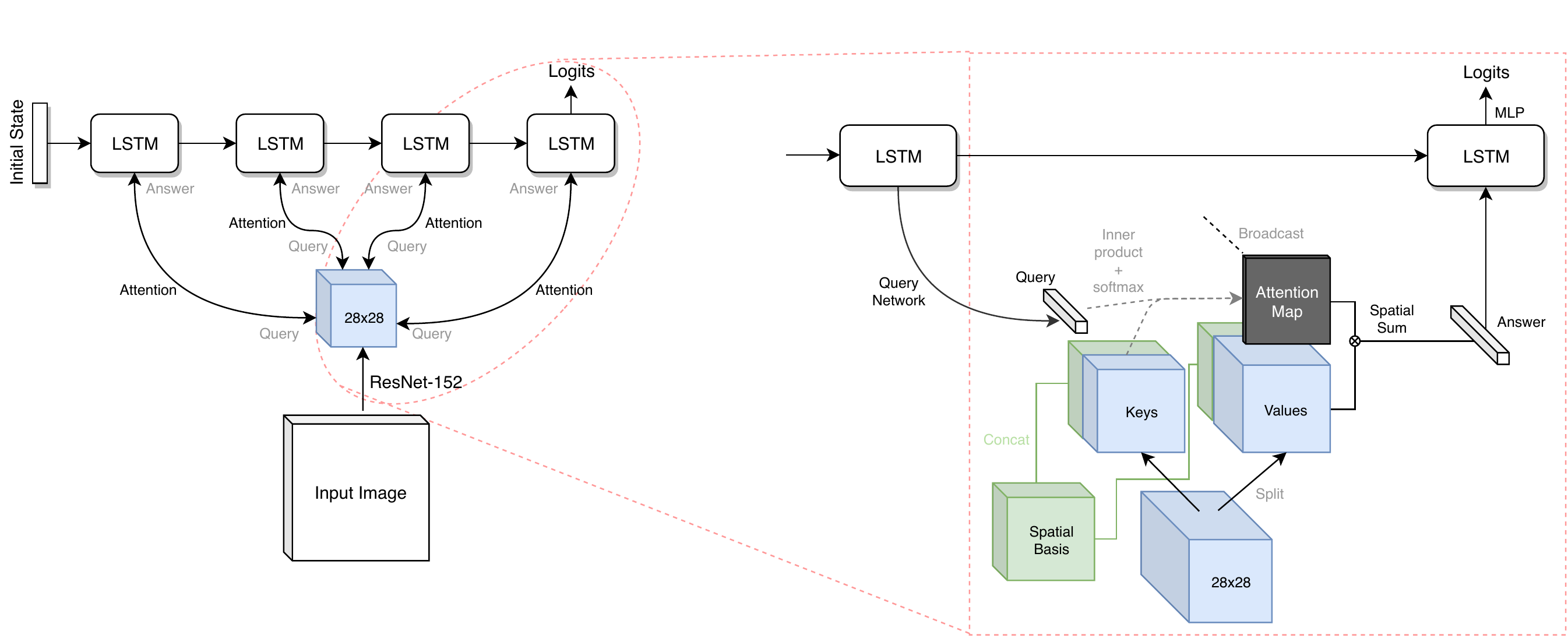}
    \caption{Full model details. Left - high level overview of the recurrent mechanism (for a 4-step model) and the corresponding vision net. Right - a zoomed in detail on one attention step. See text for full details.}
    \label{fig:full_model}
\end{figure*}
\subsection{Detailed model overview}
Here we detail and expand on the specifics of the model described in the main paper. For a more thorough overview and discussion, we refer the reader to \citet{mott2019towards}. 

Figure \ref{fig:full_model} depicts the model in full detail. On the left is the unrolled compute graph for a 4 step model. On the right is a detailed overview of the attention mechanism for a single attention head.

The model is comprised of two main components - a ``vision net'' component which processes the input image and outputs a spatial tensor (of size $28\times28$ in this case) and a ``controller'' component, which is a recurrent core attending this tensor at every time step. On the left of Figure \ref{fig:full_model} we can see the image being processed through the vision net (ResNet-152, in this case) and the output tensor being produced. This happens only once per image since the input image is fixed for all compute steps. At the top we see the controller (in this case an LSTM) unrolled. A learned initial state is used for the first time step. At each time step the model attends the vision nets' output tensor, sending out a query vector(s) and receiving an answer vector(s) to process. At the final step we read out the controller's state, pass it through an MLP and produce the class logits.

The right side of Figure \ref{fig:full_model} depicts the inner workings of the attention mechanism. The output tensor from the vision net is split into two tensors along the channel dimension. The first $C_k$ channels ($C_k=32$ in all our experiments) are used as the ``keys tensor'' (of size $28\times28\times C_k$). The rest of the channels are used as the ``values'' tensor with $C_v$ channels ($C_v=2016$ in all experiments, tensor size is $28\times28\times C_v$). To both the keys and values tensors we concatenate a ``spatial basis'' tensor of size $28\times28\times C_s$. This fixed tensor allows the attention to preserve and query about spatial information after the spatial structure is lost (see below).

We produce a ``query'' vector by passing the previous controller state (together with the last time step's logits) through a ``query network'' (an MLP). This query vector is of size $1\times 1\times (C_k+C_s)$. We take the inner product of the query vector with each \emph{spatial} position in the keys tensor to produce a $28\times 28$ single channel attention logits map. We pass this logits map through a spatial softmax to produce the final attention map for this head.

We broadcast the attention map along the channel dimension and point-wise multiply it with the values tensor. The result is then reduce summed along both spatial dimensions to produce the ``answer'' vector of size $1\times1\times(C_v+C_s)$. The importance of the spatial basis becomes clear at this point, since the summation causes all spatial structure to be lost. 

We concatenate all answer vectors together with all query vectors and use these as inputs to the controller at the current time step. We pass the controller state through an MLP to produce the class logits. If it is the last time step, we produce the final class logits and the loss is calculated from them.

\subsection{Model parameters}
We now give the full details of model parameters. Unless otherwise noted all non-linear activations are ReLUs. All learned parameters are initialized randomly.

\subsubsection{Vision net}
For the vision net we use a standard ResNet-152 V2. We change the stride to 1 in all residual blocks other than the second one and we don't use the final linear layer. The output of this network for a $224\times224$ image is a $28\times28\times2048$ tensor.

\subsubsection{Controller}
The controller is a standard LSTM with a hidden size of 1024 units. The initial state is randomly initialized and learned with the rest of the network.

\subsubsection{Query network}
The query network is a standard MLP with one hidden layer of size 1024. Its input is the same size of the controller state plus the size of logits (2024 in our case). Its output size is $(C_k+C_s)\times N$ where N=4 is the number of heads we use. The spatial basis has 64 channels, hence the total output size is 384 (with 32 key channels).

\subsubsection{Output MLP}
This is a single hidden layer MLP with 1024 units in the hidden layer. Its input is the hidden state of the controller and its output is 1000 units large (i.e. number of classes).

\subsubsection{Spatial Basis}
The spatial basis is a fixed tensor of size $28\times28\times C_s$. We use spatial sine and cosine functions to form the basis, with 4 frequencies per dimension resulting in $C_s=(2\times4)^2=64$ channels.

\section{Training and evaluation details}

We follow the training and evaluation protocol of \citet{fairdenoising2018} in most parameters. We use a gradient descent optimizer with momentum. We use a lower learning rate of $0.5e^{-1}$ scaled by a batch size of 256. For our models we use batch size of 1024 so the initial learning rate is $0.2$. We warm start for the first 5 epochs and then anneal the learning rate (with a factor of $1e^{-1}$) after 35 epochs, 70 epochs and 95 epochs. We train for 120 epochs.

The attack during training is random targeted PGD with signed gradients. We use $\epsilon=16$, random initialization probability of 0.8 and step size of 1/255. The loss function for adversarial training is only on adversarial examples, there is no loss coming from clean images. We add $L_2$ weight decay loss with weight $1e-4$, and label smoothing of 0.1. We use the same attack in evaluation, with step size of $1/255$ for all number of steps other than 10 where we use $1.6/255$, again following \citet{fairdenoising2018}.

\section{Additional results}
In Table \ref{tab:extra_strong} we report results on extra strong attacks with 5000 and 10,000 steps --- due to the cost of evaluation we evaluated only on 2200 random images from the ImageNet test set, but in our experience numbers change very little on larger evaluation sets.

\begin{table*}[]
\centering
\begin{tabular}{l|c|c|c|c|}
\cline{2-5}
\multicolumn{1}{c|}{\textbf{}} & \multicolumn{2}{c|}{5000 steps} & \multicolumn{2}{c|}{10,000 steps} \\ \cline{2-5} 
\textbf{Model} & \multicolumn{1}{c|}{\begin{tabular}[c]{@{}c@{}}Top-1\\ Accuracy\end{tabular}} & \multicolumn{1}{c|}{\begin{tabular}[c]{@{}c@{}}Attack\\ Success Rate\end{tabular}} & \multicolumn{1}{c|}{\begin{tabular}[c]{@{}c@{}}Top-1\\ Accuracy\end{tabular}} & \multicolumn{1}{c|}{\begin{tabular}[c]{@{}c@{}}Attack\\ Success Rate\end{tabular}} \\ \hline
\textbf{ResNet-152} & 28.12\% & 58.03\% & 27.64\% & 58.79\% \\
\textbf{S3TA-2} & 32.85\% & 52.50\% & 32.63\% & 52.94\% \\
\textbf{S3TA-4} & 35.67\% & 50.57\% & 35.43\% & 50.96\% \\
\textbf{S3TA-8} & 36.20\% & 50.75\% & 35.44\% & 51.91\% \\
\textbf{S3TA-16} & 37.36\% & 49.50\% & 36.96\% & 50.23\% \\ \hline
\textbf{ResNet-152-30} & 36.69\% & 41.33\% & 36.78\% & 41.25\% \\
\textbf{DENOISE} & 42.76\% & 33.83\% & 42.99\% & 33.88\% \\
\textbf{S3TA-16-30} & \textbf{46.11\%} & \textbf{24.91\%} & \textbf{46.20\%} & \textbf{24.68\%}
\end{tabular}
\caption{Results for extra strong PGD attacks with 5000 and 10000 steps. The general picture remains as S3TA-16-30 is significanly more robust than all other models. S3TA-16 trained with 10 PGD steps is more accurate than a ResNet-152 trained with 30 PGD steps even under these powerful attacks.}
\label{tab:extra_strong}
\end{table*}

\subsection{PGD attacks with multiple restarts}
Here we test the three strongest models against attacks with multiple restarts. We test with the same PGD with signed gradient attack used so far (with a 100 and 1000 steps) but we allow the attacker to start from multiple random initialization: 1, 10 or 100. We take the strongest example from all restarts to measure the top-1 and attack success rate (i.e. even if one of the adversarial attacks succeeds we count it as a success, and if the network mislabels a single image out of the attacks we do not increase the top-1 score). Results are reported in Table \ref{tab:restarts}. As can be seen, all models are quite robust to multiple restarts, and S3TA-16-30 is better than the other models.

\begin{table*}[]
\centering
\begin{tabular}{l|c|c|c|c|c|c}
 Attack steps & \multicolumn{3}{l|}{100 steps} & \multicolumn{3}{l|}{1000 steps} \\ \cline{1-7}
 Random restarts & 1 & 10 & 100 & 1 & 10 & 100 \\ \cline{1-7}
\textbf{ResNet-152-30} & 39.77\% & 39.37\% & 39.15\% & 37.41\% & 36.51\% & 35.75\%  \\
\textbf{DENOISE} & 46.16\% & 44.68\% & 44.68\% & 43.57\% & 41.91\% & 41.37\% \\
\textbf{S3TA-16-30} & \textbf{47.76\%} & \textbf{47.27\%} & \textbf{47.09\%} & \textbf{46.74\%} & \textbf{46.09\%} & \textbf{45.78\%} \\
\end{tabular}
\caption{PGD attack with multiple restarts. We test the various method under attacks starting at multiple random initialization points. We test with 100 and 1000 PGD attack steps, initialized at 1, 10 and 100 random starting points. We count the attack successful even if only one of the adversarial examples caused the model to output the target class. We only count an image to be correctly classified if it was correctly classified across all adversarial images. As can be see, all models are quite robust here, though again, S3TA-16-30 outperforms them all.}
\label{tab:restarts}
\end{table*}

\subsection{Random targeted attacks with an Adam optimizer}
It has been observed \citet{gowal2019alternative} that random targeted attacks using the FGSM method are sometimes weaker than attacks optimized with the Adam optimizer. Here we test the performance of the model with this potentially stronger type of targeted attack. We use a 250 step attack, optimized with the Adam optimizer with learning rate annealing (0.1 until step 100, 0.01 until 200 and 0.001 for the last 50). This has been shown \citet{gowal2019alternative} to be quite a strong attack. Results are reported in Table \ref{tab:adam}. As can be seen all models perform worse here than with PGD with iterative FGSM, however S3TA-16-30 is significantly more robust here as well.

\begin{table*}[]
\centering
\begin{tabular}{@{}ccc@{}}
\toprule
 & \textbf{Top-1} & \textbf{Success rate} \\ \midrule
\textbf{ResNet-152-30} & 26.38\% &42.94\% \\ \midrule
\textbf{DENOISE} & 32.99\% &35.04\% \\ \midrule
\textbf{S3TA-16-30} & \textbf{36.83\%} &\textbf{27.00\%} \\ \midrule
\end{tabular}
\caption{Random targeted PGD attack with Adam optimizer. We measure the robustness of some models under an random targeted attack optimized with the Adam optimizer (rather than iterative FGSM). The attack uses 250 iterations, with learning rate 0.1 for the first 100 iterations, 0.01 for next 100 and 0.001 for the last 50. This has be shown \citet{} to be a stronger attack the iterative FGSM. As can S3TA-16-30 is significantly better at defending here compared to the other models.}
\label{tab:adam}
\end{table*}

\section{More adversarial examples}
\subsection{Non visible examples}
\begin{figure*}[h]
    \centering
    \includegraphics[width=0.7\textwidth]{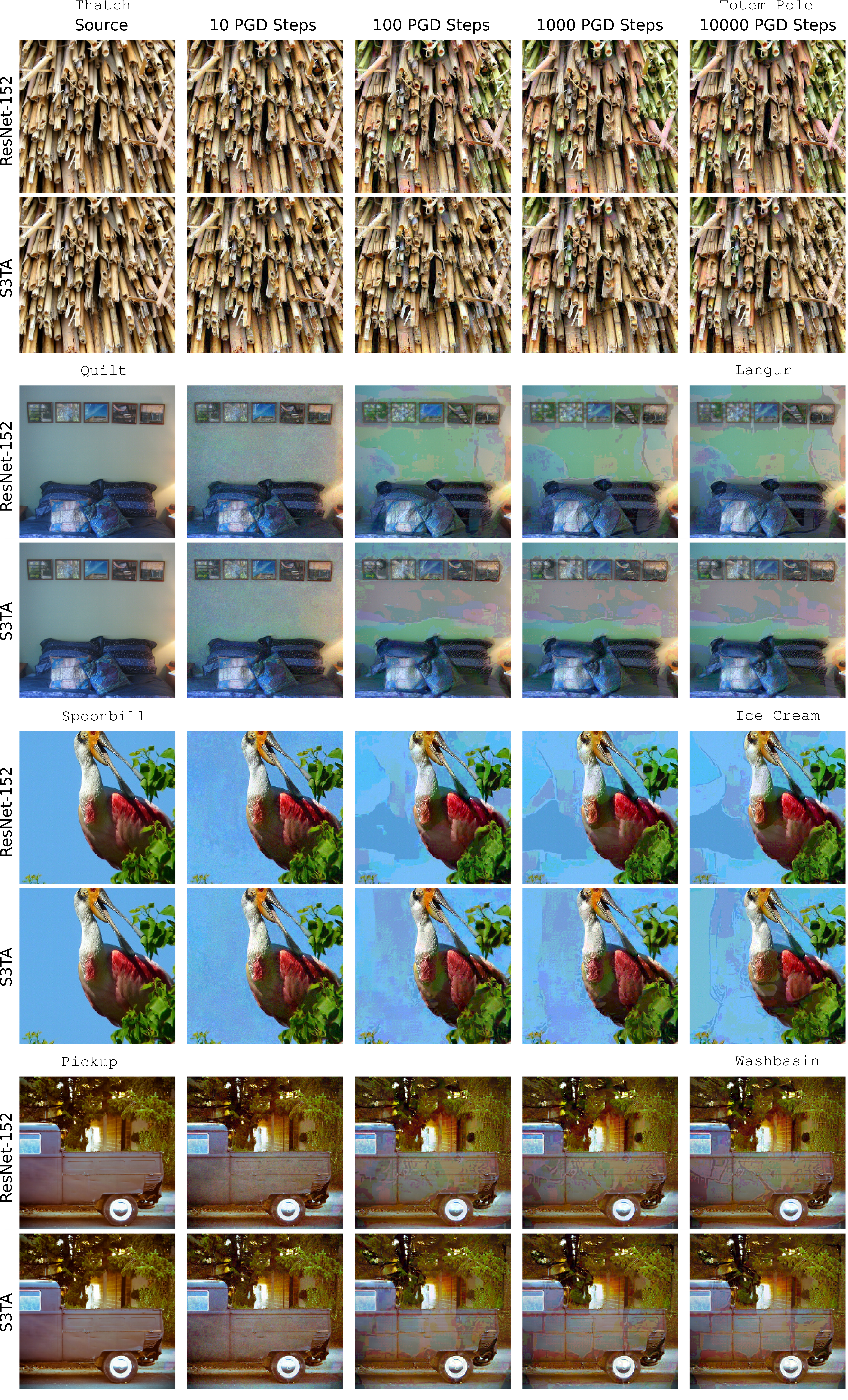}\vspace{-1.0em}
    \caption{Adversarial images generated by PGD to attack the ResNet-152 and S3TA-4 model with no apparent visual structure. The source label is depicted on the left, the target label on the right. Source images are the leftmost column. We generate examples for 10, 100, 1000 and 10,000 PGD steps. These are cases where the perturbation does not have clear structure, as for many of the images. Some local structure may be interpretable - tiles for the washbasin for example, but there's little coherent global structure. Images are best viewed on screen and zoomed in.}
    \label{fig:bad_examples}
\end{figure*}

Figure \ref{fig:bad_examples} shows adversarial examples for a S3TA-4 model where there is no visible structure to be seen, as is often the case with fully trained models.
\subsection{Structured examples - mid training}
\begin{figure*}[h]
    \centering
    \includegraphics[width=0.7\textwidth]{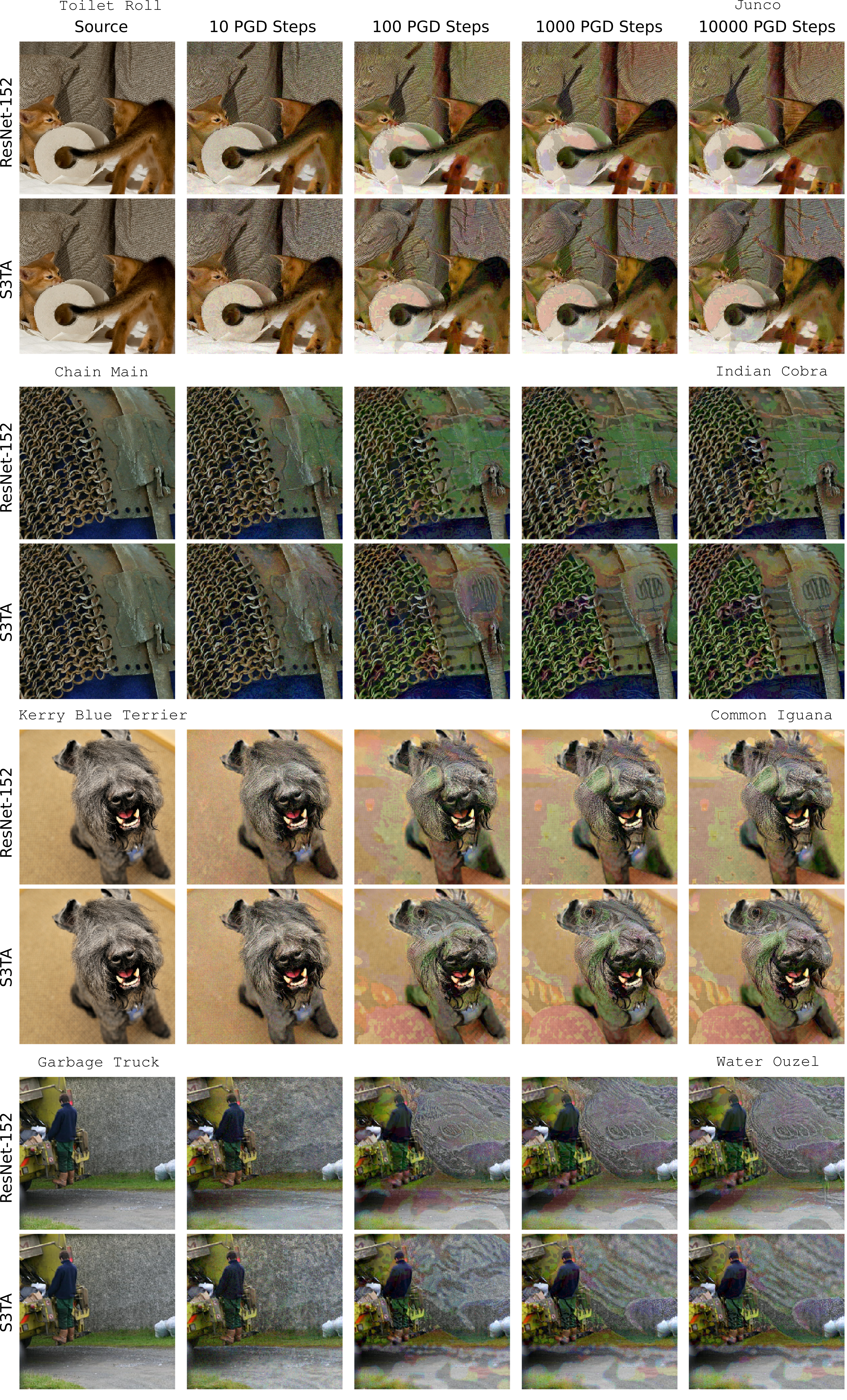}\vspace{-1.0em}
    \caption{Adversarial images generated by PGD to attack the ResNet-152 and S3TA-4 model with clear and intrepretable structure. The source label is depicted on the left, the target label on the right. Source images are the leftmost column. We generate examples for 10, 100, 1000 and 10,000 PGD steps. While the ResNet attacks are not particularly interesting, the S3TA examples are extremely structured. Note the salient, global and often realistic structures appearing - the Junco bird, the head of the Iguana, the body of the Indian Cobra or the Water Ouzel, a bird which is often depicted next to water (which is also created). We encourage the reader to view these on screen and zoomed-in.}
    \label{fig:mid_examples}
\end{figure*}

Figure \ref{fig:mid_examples} shows adversarial examples for a S3TA-16 model around half-way through training, with clear salient global structures. These are much more common before training concludes (but still appear in fully trained models as in the main paper). Understanding the exact circumstances under which these appear is an open question.

\end{document}